\documentclass{article}
\usepackage{algorithmic}
\usepackage{enumitem}

\usepackage{microtype}
\usepackage{graphicx}
\usepackage{subcaption}
\usepackage{booktabs} 
\usepackage{natbib}

\usepackage{hyperref}

\usepackage[preprint]{icml2026}

\usepackage{amsmath}
\usepackage{amssymb}
\usepackage{mathtools}
\usepackage{amsthm}
\usepackage{multirow}

\usepackage[capitalize,noabbrev]{cleveref}

\theoremstyle{plain}

\theoremstyle{definition}

\theoremstyle{remark}

\usepackage[textsize=tiny]{todonotes}

\icmltitlerunning{Path-Guided Flow Matching for Dataset Distillation}

\begin{document}

\twocolumn[
  \icmltitle{Path-Guided Flow Matching for Dataset Distillation}



  \icmlsetsymbol{equal}{*}

  \begin{icmlauthorlist}
    \icmlauthor{Xuhui Li}{ml}
    \icmlauthor{Zhengquan Luo}{ml}
    \icmlauthor{Xiwei Liu }{cv}
    \icmlauthor{Yongqiang Yu}{cv}
    \icmlauthor{Zhiqiang Xu}{ml}
  \end{icmlauthorlist}


  \icmlaffiliation{ml}{Department of Machine Learning, MBZUAI, Abu Dhabi, UAE}
  \icmlaffiliation{cv}{Department of Computer Vision, MBZUAI, Abu Dhabi, UAE}
  \icmlcorrespondingauthor{Zhiqiang Xu}{zhiqiang.xu@mbzuai.ac.ae}
  \icmlkeywords{Machine Learning, ICML}

  \vskip 0.3in
]



\printAffiliationsAndNotice{}  

\begin{abstract}
Dataset distillation compresses large datasets into compact synthetic sets with comparable performance in training models.
Despite recent progress on diffusion-based distillation, this type of method typically depends on heuristic guidance or prototype assignment, which comes with time-consuming sampling and trajectory instability and thus hurts downstream generalization especially under strong control or low IPC. 
We propose \emph{Path-Guided Flow Matching (PGFM)}, the first flow matching-based framework for generative distillation, which enables fast deterministic synthesis by solving an ODE in a few steps. PGFM conducts flow matching in the latent space of a frozen VAE to learn class-conditional transport from Gaussian noise to data distribution. 
Particularly, we develop a continuous path-to-prototype guidance algorithm for ODE-consistent path control, which allows trajectories to reliably land on assigned prototypes while preserving diversity and efficiency.
Extensive experiments across high-resolution benchmarks demonstrate that PGFM matches or surpasses prior diffusion-based distillation approaches with fewer steps of sampling while delivering competitive performance 
with remarkably improved efficiency, e.g., 7.6$\times$ more efficient than the diffusion-based counterparts with 78\% mode coverage.
\end{abstract}



\section{Introduction}
The rapid progress of machine learning has been driven by larger datasets and models, but this trend also makes training increasingly expensive in compute and storage. Beyond model-side efficiency techniques such as pruning \citep{liu2017learning, ding2019centripetal} and quantization \citep{wu2016quantized, chen2021towards, chauhan2023post}, an important alternative is to reduce the data required for training. Dataset distillation \cite{wang2018dataset, liu2022dataset} does it by compressing a large training dataset into only a few synthetic samples per class (IPC), such that a model trained on the distilled dataset approaches the performance of the model trained on the full dataset. Despite the importance and popularity, an inherent difficulty in distilling data lies in that synthesized datasets must remain class-discriminative while capturing intra-class diversity, which is particularly pronounced at low IPC, where each sample must represent multiple modes without collapsing to class averages. On the other hand, as IPC grows, the trade-off between mode coverage and trainable details of images must be balanced within class.

Diffusion-based generative models have emerged as a dominant approach for dataset distillation, with methods such as DiT \cite{peebles2023scalable}, MinimaxDiff \cite{gu2024efficient}, and MGD$^3$ \cite{chan2025mgd} achieving strong performance especially in the low-IPC regime. However, such models typically inherit the complexity of diffusion sampling resulting from carefully designed noise schedules \cite{ho2020DenoisingDiffusionProbabilistic}, classifier-free guidance (CFG) \cite{ho2022classifier}, or additional guidance heuristics, which increases sampling cost and causes instability. This raises a natural question: \emph{Is there any more efficient, and more stable generative paradigm that can distill comparable or even better datasets?} 

\begin{figure}[t]
    \centering
    \includegraphics[width=\linewidth]{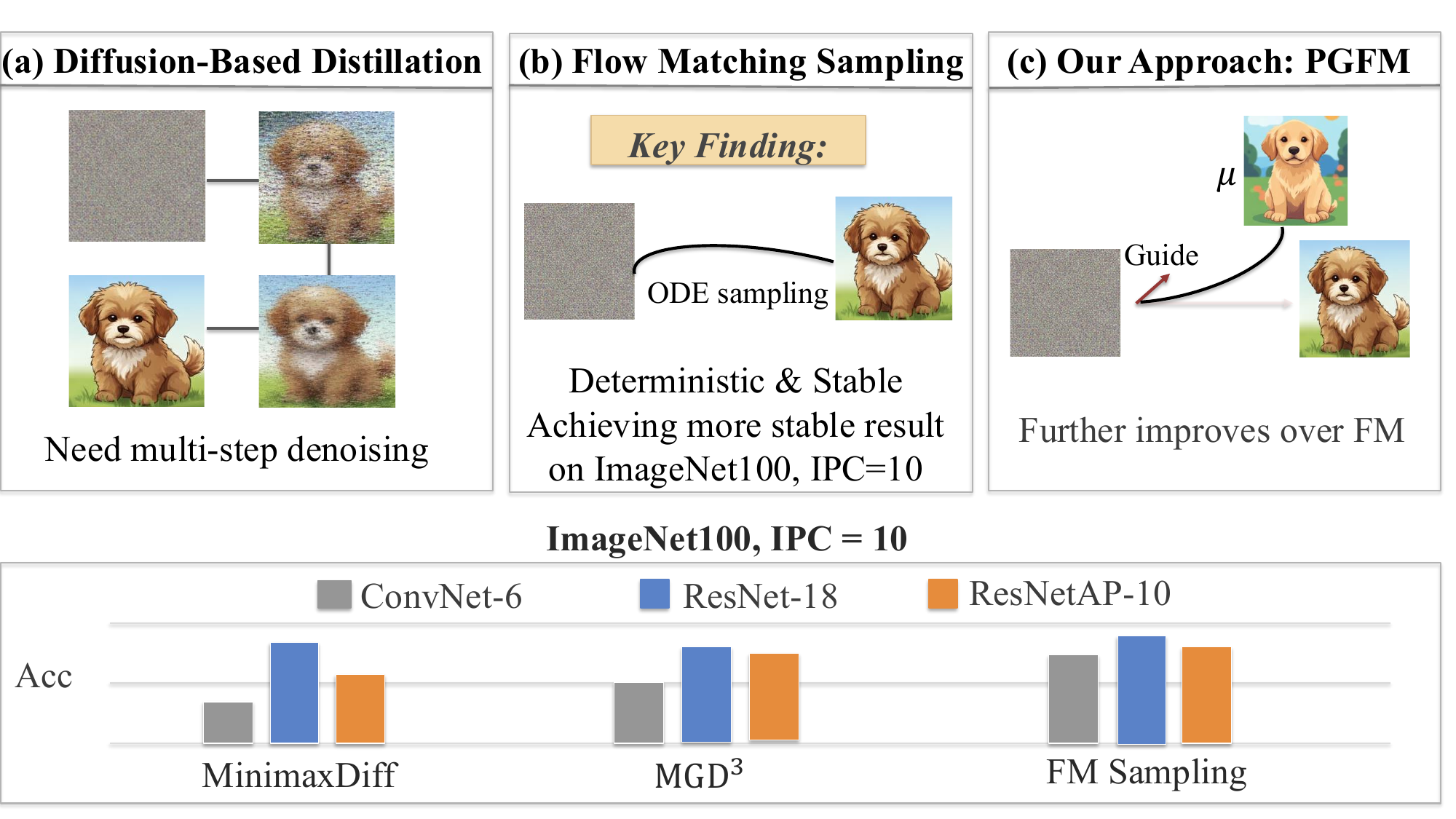}
    \caption{
    \textbf{Motivation:}
    Diffusion sampling denoises step-by-step, while flow matching uses deterministic ODE sampling that is already strong and smooth; PGFM adds lightweight prototype guidance to further improve performance.
    }
    \label{fig:intro}
\end{figure}

Flow matching \cite{lipman2022flow, lipman2024flow} has recently been rising as an alternative generative paradigm to diffusion models, which learns a time-dependent velocity field and generates samples by integrating a deterministic (or lightly stochastic) flow/ODE \cite{hartman2002ordinary} from a simple base distribution to the data distribution. Compared to diffusion models that rely on repeated denoising with explicit noise schedules and often strong CFG, flow-matching sampling is conceptually cleaner and can be more stable and efficient in practice \cite{dao2023flow, chen2023flow}. These properties are particularly appealing for dataset distillation, where distilled datasets must be both representative (stay close to the class manifold) \cite{li2025geodm} and diverse (cover multiple modes) under tight budgets. Empirically, we find that on ImageNet100 with IPC at 10, even pure flow-matching sampling without any additional guidance already outperforms strong diffusion-based distillation baselines such as MGD$^3$ and MinimaxDiff in Figure~\ref{fig:intro}. This suggests that the flow-matching trajectory may be inherently well-suited for producing “trainable” distilled samples. However, the challenges on how to encourage diversity coverage and feature alignment for classification remain to be overcome by flow matching.

To this end, we propose the first flow-matching method for dataset distillation, i.e., Path-Guided Flow Matching or \textbf{PGFM} for short. We use a pretrained \textbf{G}aussian \textbf{M}ixture \textbf{Flow} matching (GMFlow) \cite{chen2025gaussian} generator as the backbone and generate synthetic images via standard flow sampling. To encourage diversity coverage, we first perform per-class mode discovery in latent space, and assign one prototype to each synthesized sample. A key empirical finding is that overly strong control can easily over-regularize the trajectory, pulling samples toward blurry or even noisy-looking images (see Appendix \ref{append:bulr}). Therefore, PGFM exerts only little prototype-guided control over sampling, together with early stopping and a simple trust-region constraint to limit the control strength, in the initial sampling phase. For the later sampling phase, the flow model is aimed at recovering fine-grained details. Overall, PGFM stays fully compatible with flow matching and deliberately avoids heavy heuristics, as a clean and interpretable way to incorporate mode guidance into stable flow-based generation.

Empirically, PGFM delivers consistent accuracy improvements with substantially higher efficiency across high-resolution datasets. On ImageNette, PGFM reliably outperforms the strongest baseline, i.e., MGD$^3$, at low budgets like $\text{IPC}=10$ or $20$, while the improving margin naturally shrinks as IPC increases. On the harder ImageNet100 benchmark, we find that unguided flow-matching sampling is already a surprisingly strong baseline, and PGFM further boosts performance and performs best across evaluation backbones. At the same time, PGFM improves synthesis efficiency by 7.6 $\times$ over the diffusion-based state-of-the-art method and achieves a $78\%$ prototype hit rate, meaning a dramatically higher level of mode coverage. Overall, the experimental study indicates that flow matching with lightweight and early-stage prototype guidance can generate distilled datasets with better performance and model coverage at a significantly lower cost, especially in tight-budget and many-class settings.

We make the following key contributions in this work:
\begin{itemize}
  \item[$\bullet$] We are the first to propose flow matching for dataset distillation as an efficient alternative to existing generative distillation paradigms. Particularly, we propose path-guided flow matching (PGFM).
  \item[$\bullet$] To enhance distribution coverage and performance especially in scenarios of large IPC, we introduce a prototype-guided strategy which enables the flow matching model to better capture and represent diverse intra-class modes.
  \item[$\bullet$] Extensive experiments across high-resolution benchmarks show that PGFM improves the accuracy over state-of-the-art methods while significantly reducing computational costs.
\end{itemize}

\section{Related Work}

\textbf{Optimization-based dataset distillation.}
Early dataset distillation methods optimize a small synthetic set by matching training signals from real data. Gradient matching \cite{zhao2021dataset} aligns gradients between real and synthetic batches, but relies on expensive bi-level optimization and scales poorly. To improve efficiency, Distribution Matching (DM) \cite{zhao2023dataset} matches feature statistics under randomly initialized networks, avoiding inner-loop training. Trajectory-based approaches further strengthen the signal by matching multi-step training dynamics, e.g., MTT \cite{cazenavette2022distillation}. M3D \cite{zhang2024m3d} formulates condensation into a min--max problem that reduces worst-case mismatch, while DSDM \cite{li2024diversified} matches semantic distributions with explicit diversity regularization. More recently, NCFM \cite{wang2025dataset} proposes a neural characteristic-function discrepancy from a min-max perspective to improve robustness.


\textbf{Generative dataset distillation.}
Generative dataset distillation trains a generator to synthesize the entire distilled dataset, rather than optimizing each synthetic image individually \cite{zhang2023dataset, gu2024efficient, su2024d}. Early work \cite{zhang2023dataset} explored class-conditional GANs with learnable discrete latent representations and auxiliary losses for realism, representativeness, and diversity. More recent methods largely adopt diffusion models as the backbone due to their high-fidelity synthesis. Representative diffusion-based approaches include MinimaxDiffusion \cite{gu2024efficient}, which fine-tunes a pretrained model with distribution-level objectives; $D^4$M \cite{su2024d}, which performs mode-aware sampling by injecting structured noisy modes; Influence-guided Diffusion (IGD) \cite{chen2025influence}, which uses influence estimates to guide sampling toward more training-beneficial samples; and MGD$^3$ \cite{chan2025mgd}, which discovers class-conditional modes (e.g., via latent clustering) and applies mode guidance to improve coverage under tight budgets.

\textbf{Flow matching strategy.}
Flow Matching (FM) and Rectified Flows \cite{lipman2022flow, liu2022rectified} offer an alternative generative paradigm that learns a continuous velocity field transporting a simple base distribution to the data distribution, with sampling formulated as ODE integration.
In practice, this can substantially reduce sampling steps (e.g., 5--10) and provides a cleaner interface for incorporating lightweight controls.
These properties make FM a promising foundation for generative dataset distillation, motivating our flow-matching based approach that achieves strong distilled performance while avoiding the heavy sampling cost and heuristic complexity commonly associated with diffusion-based pipelines.

\section{Preliminaries}

\subsection{Dataset Distillation}
Given a large-scale dataset with training set $\mathcal{U} = \{(x_i, y_i)\}_{i=1}^{N_{\mathcal{U}}}$, the goal of dataset distillation (DD) is to build a smaller synthetic dataset $\mathcal{S} = \{(\tilde{x}_i, \tilde{y}_i)\}_{i=1}^{N_{\mathcal{S}}}$, where $N_{\mathcal{S}} \ll N_\mathcal{U}$ and $x_{i}, \tilde{x}_{i}$ are the original and synthetic samples\footnote{Throughout the paper, samples are images.} with corresponding class labels $y_i, \tilde{y}_i$. The essential requirement of DD lies in that the model $\phi_{\mathcal{S}}$ trained on the smaller synthetic dataset $\mathcal{S}$ should achieve similar test performance to that of a model $\phi_{\mathcal{U}}$ trained on the original set $\mathcal{U}$. That is, if $\mathcal{A}$ is the accuracy of a model on the test set $\mathcal{U}_{e}$, then $\mathcal{A}(\phi_{\mathcal{U}}) \sim \mathcal{A}(\phi_{\mathcal{S}})$. The size of the distilled dataset $N_{\mathcal{S}}$ is defined by the distillation budget, denoted as IPC (images per class). Prior generative models \cite{gu2024efficient, su2024d, zhang2023dataset} conduct dataset distillation by approximating the dataset distribution through sampling diverse and representative instances. This line of research on \emph{generative dataset distillation} tries to minimize the distribution gap which is often characterized by the following loss matching criterion:
\begin{equation*}
\Big\| \mathbb{E}_{x \sim P_{data}}\big[ \ell (\phi_{\mathcal{U}}(x), y) \big] - \mathbb{E}_{x \sim P_{syn}}\big[ \ell (\phi_{\mathcal{S}}(\tilde{x}), y) \big] \Big\| < \epsilon,
\end{equation*}
where $P_{[\cdot]}$ is the underlying data distribution, $\epsilon$ is the error bound constant, and $\ell$ is a loss function. The use of generative models provides flexibility and privacy protection by not being limited to selecting original samples.

\subsection{Latent Space Generative Modeling}
To enable dataset distillation at higher resolutions, it is common to operate in the latent space of a pre-trained Variational Autoencoder (VAE) with a frozen encoder $E$ and a frozen decoder $D$ (e.g., adopted from Stable Diffusion). The encoder induces a perceptual latent space $\mathcal{Z}$ by mapping each image $x$ to a posterior distribution $q_E(z\mid x)$. This defines an aggregated latent distribution over the dataset, as well as its class-conditional variant:
\[
\begin{aligned}
& q_E(z) = \mathbb{E}_{x\sim P_{\text{data}}}\!\left[q_E(z\mid x)\right],
\\
& q_E(z\mid y) = \mathbb{E}_{x\sim P_{\text{data}}(\cdot\mid y)}\!\left[q_E(z\mid x)\right].
\end{aligned}
\]
In practice, the posterior mean is often used as a deterministic latent representation,
\[
\bar z(x) = \mathbb{E}_{q_E(z\mid x)}[z],
\]
which yields, for each class $y$, an empirical latent set
\[
\mathcal{Z}_y = \{\bar z(x_i)\,:\, y_i = y\},
\]
providing a compact description of the class-conditional data distribution in latent space and allowing reconstruction back to the image space via the frozen decoder $D$.

\subsection{Conditional Flow Matching (CFM) and ODE Sampling}
Flow matching trains a continuous generative model by learning a conditional velocity field $u_\phi(t,z,y)$ that transports a simple distribution to the class-conditional data distribution along a continuous time path $t\in[0,1]$. A common training objective is to regress the model velocity to a target velocity field $u^\star$ constructed from a coupling between $z_0\sim p_0$ and $z_1\sim p_1(\cdot\mid y)$:
$$
\mathcal{L}(\phi)=\mathbb{E}_{y,\;t,\;z_t}\Big[\lambda(t)\,\big\|u_\phi(t,z_t,y)-u^\star(t,z_t,y)\big\|_2^2\Big].
$$
After training, sampling boils down to solving an ordinary differential equation (ODE) from noise to data.

In this work, we do not train a flow model from scratch. Instead, we adopt a pre-trained flow-matching generator, i.e., GMFlow, to parameterize the class-conditional velocity field $u_\phi(t,z,y)$ in the VAE latent space. Given a class label $y$, we start from a base latent
$
z(0)=z_0, 
$
where $z_0 \sim \mathcal{N}(\mathbf{0},\mathbf{I})$, 
and generate a synthetic latent by integrating
\begin{equation}
\label{eq:ode}
\frac{dz(t)}{dt} = u_\phi(t,z(t),y), \qquad t\in[0,1],
\end{equation}
to obtain $\tilde z = z(1)$. The synthetic image is then recovered via the frozen VAE decoder,
$
\tilde X = D(\tilde z).
$
In practice, we solve the ODE with efficient numerical solvers (e.g., Heun's method) using a small number of steps. Our proposed guidance is only applied during this sampling process, without modifying the pre-trained generator.

\section{Proposed Method: PGFM}

\begin{figure}[t]
    \centering
    \includegraphics[width=\linewidth]{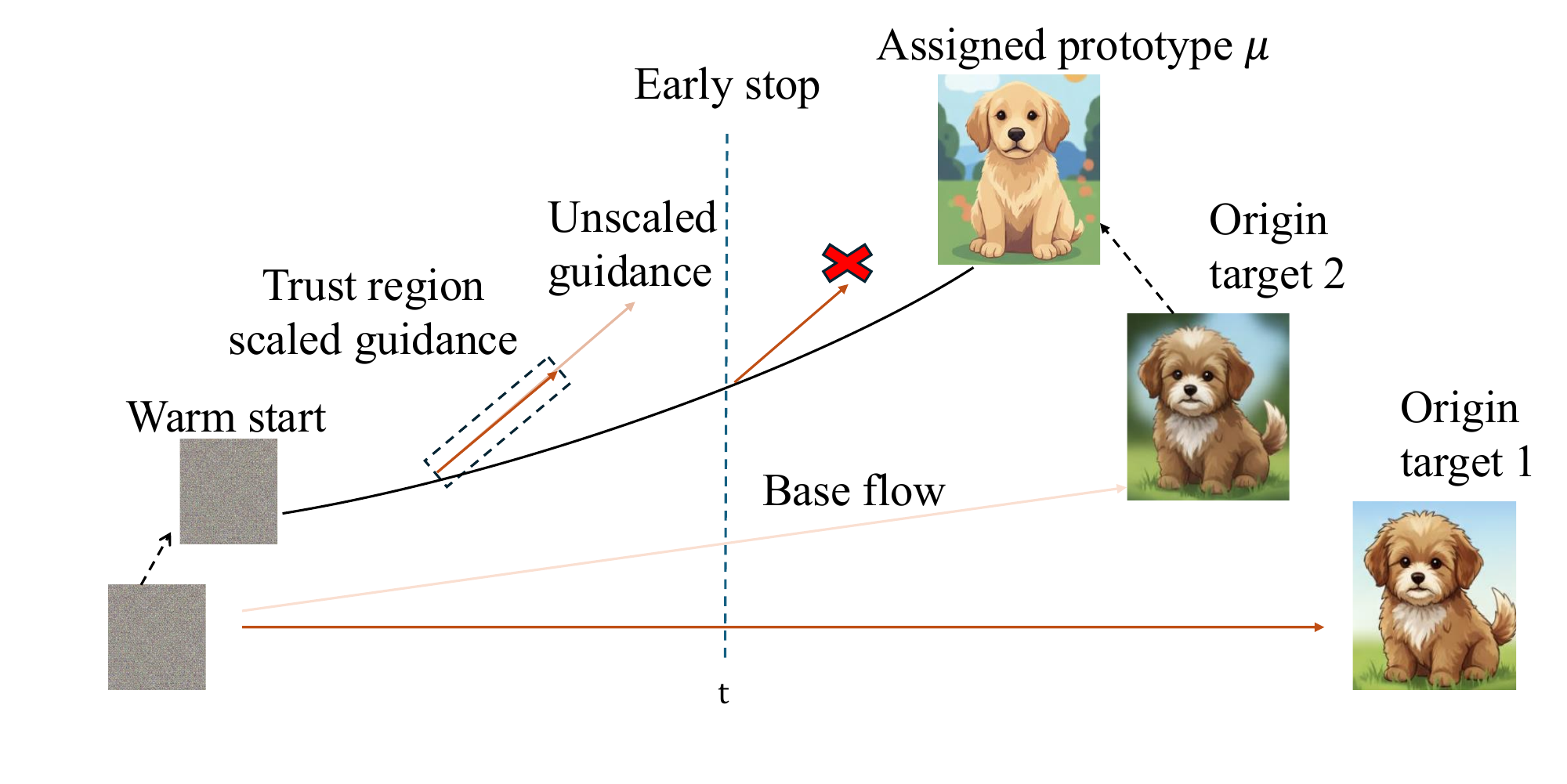}
    \caption{
    \textbf{Sampling process of PGFM:}
    Starting from Gaussian noise, we then sample with a pretrained flow-matching generator (GMFlow) while applying lightweight, early-stage prototype guidance with a trust region to improve mode coverage without washing out details.
    }
    \label{fig:overview}
\end{figure}

Our proposed Path-Guided Flow Matching (PGFM) builds on a pre-trained flow-matching generative model in a compressed VAE latent space to synthesize diverse and representative data efficiently. 
To improve mode coverage under tight budgets, we further introduce a lightweight prototype guidance mechanism which leverages early-stage sampling trajectories for class-specific latent prototypes. 

The overall PGFM pipeline consists of i) latent-space preprocessing, ii) flow-matching sampling in latent space, and iii) latent selection for decoding, to obtain the final synthetic dataset. Figure~\ref{fig:overview} describes the flow matching sampling process of which the details can be found in Algorithm \ref{alg:pgfm}, Appendix \ref{append:alg}.

\subsection{Latent Space Preprocessing}
We start by moving dataset distillation from pixel space to a compact VAE latent space, which is more efficient and perceptually aligned.
We therefore use a frozen, pre-trained VAE \cite{doersch2016tutorial} with encoder $E$ and decoder $D$, 
which enables distillation at higher resolutions while keeping synthesis computationally light. 
Given the training dataset $\mathcal{U}=\{(x_i,y_i)\}_{i=1}^{N_{\mathcal{U}}}$, we encode each image into a latent representation 
$z_i = \mathrm{mean}(E(x_i))$, 
and obtain a latent dataset $\mathcal{V}=\{(z_i,y_i)\}_{i=1}^{N_{\mathcal{U}}}$. 
For each class $y$, we collect the empirical latent pool
$
\mathcal{Z}_y = \{z_i \,:\, y_i = y\},
$
which serves as a discrete approximation to the class-conditional target distribution $p_1(z\mid y)$ in latent space. 
In practice, we keep $E$ and $D$ frozen throughout and perform all prototype construction and flow sampling in this latent space. 
For numerical stability, we standardize latent data using their statistics: given all encoded latent data $\{ z_i\}$, we compute the global mean $\mu$ and per-dimension standard deviation $\sigma$, and apply
$\tilde z = ( z - \mu)/\sigma$.

To encourage broad coverage of the class-conditional latent space under a tight budget, we further perform $K$-means++ clustering \cite{bahmani2012scalable}
on each $\mathcal{Z}_y$ with $K=\mathrm{IPC}$ and use the resulting cluster centers 
$\{\mu_{y,k}\}_{k=1}^{K}$ as class-specific prototypes, which serve as target modes for the guided sampling stage described next.

\subsection{Flow-Matching Sampling}
PGFM synthesizes samples by integrating the GMFlow in the VAE latent space.
Given class label $y$, we first draw an initial latent
\[
z_0 \sim p_0(z)=\mathcal{N}(\mathbf{0},\mathbf{I}),
\]
and generate a sample by solving the latent-space ODE with Eq.~\eqref{eq:ode}
where $u_\phi$ is a class-conditional velocity field parameterized by GMFlow.
In practice, we apply an explicit ODE solver (e.g., Euler/Heun) to obtain the final latent $\tilde z=z(1)$, which is decoded into the synthetic image $\tilde x = D(\tilde z)$ using the frozen VAE decoder.

\paragraph{Path-guided control (PGFM).}
We therefore compute $K$ latent prototypes for each class, and assign each synthetic sample to a target prototype $\mu_{y,k}$.
During sampling, we augment the pretrained flow field with a lightweight prototype-guided control term:
\begin{equation}
\frac{dz(t)}{dt}
= u_\phi\!\big(t,z(t),y\big)\;+\; g(t)\,\tilde u_{\mathrm{proto}}\!\big(t,z(t),y,k\big),
\label{eq:guided_flow_ode}
\end{equation}
where $g(t)$ is an early-stop gate and the prototype control is defined by the predicted clean data \footnote{GMFlow provides an extra interface to predict the clean latent, this is not computed by the ODE solver.} $\hat{z}_1$ in the latent space:
\begin{equation}
\tilde u_{\mathrm{proto}}\!\big(t,z(t),y,k\big) \;=\; \alpha(t)\,u_{\mathrm{proto}}\!\big(t,z(t),y,k\big),
\label{eq:proto_scaled}
\end{equation}
\begin{equation}
u_{\mathrm{proto}}\!\big(t,z(t),y,k\big)
=\lambda\Big(\mu_{y,k}-\hat z_1(t)\Big),
\label{eq:proto_control}
\end{equation}
\begin{equation}
\alpha(t)
=\min\!\left(1,\; \rho_0 \frac{\big\|u_\phi(t,z(t),y)\big\|_2}{\big\|u_{\mathrm{proto}}(t,z(t),y,k)\big\|_2+c}\right).
\label{eq:trust_region}
\end{equation}
Here $\alpha(t)$ is an adaptive factor which defines our \emph{trust region} with the aim to prevent $\tilde u_{\mathrm{proto}}\!\big(t,z(t),y,k\big)$ from dominating the model dynamics in the velocity space.

This guarantees that the guided velocity satisfies
\[
\big\|g(t)\,\tilde u_{\mathrm{proto}}(t,\cdot)\big\|_2 \;\le\; \rho_0\,\big\|u_\phi(t,\cdot)\big\|_2,
\]
where $\rho_0$ controls the maximum relative strength of prototype guidance and constant $c$ avoids numerical issues. In practice, we use a time-dependent trust region for guidance and linearly decay the coefficient from $\rho_0$ to $\rho_{\min}$ until the guidance stop time $s_{\mathrm{end}}$, i.e., 
\[
\rho(t)=\rho_0 + (\rho_{\min}-\rho_0)\cdot \frac{s(t)}{s_{\mathrm{end}}}, \quad s(t)\le s_{\mathrm{end}},
\]
and guidance is disabled by $g(t)$ afterwards.

Here $\hat z_1(t)$ is the model's denoised estimate of the terminal VAE latent, computed from the current state $z(t)$ at time $t$ via the GMFlow sampling interface, and $\lambda$ controls the guidance strength.
Empirically, over-strong guidance tends to wash out fine details and yields gray or blurred samples (see Appendix \ref{append:bulr}). Hence, we introduce an \textit{early stop} strategy by a simple time gate $g(t)$, using a progress variable $s(t)\!\in\![0,1]$:
\begin{equation}
\label{eq:early_stop}
g(t)=\mathbb{I}\!\left[s(t)\le s_{\mathrm{end}}\right].
\end{equation} 
Guidance is activated only in the early trajectory, and details are recovered later by the base flow.

\paragraph{Warm start.}
For mode coverage, we apply a lightweight initialization to generate noises close to the assigned prototypes:
\begin{equation}
\label{eq:warm_start}
z_0 \leftarrow (1-\eta_{\mathrm{init}})\,z_0 + \eta_{\mathrm{init}}\,\mu_{y,k},
\end{equation}
which stabilizes early sampling and improves mode alignment under tight budgets.

\subsection{Latent Selection for Decoding}
After the flow matching sampling, the solver may present residual inconsistencies across samples, 
and different latent representations can lead to noticeably different visual quality when decoded (e.g., gray or blur artifacts).
Therefore, our implementation chooses the decoding latent between two candidates produced during the last sampling step:
(i) the final latent state $z_{1}$ (the solver state), and (ii) the final clean-latent prediction $\hat z_{1}$.
We select the most stable candidate by minimizing the batch-wise latent scattering:
\begin{equation}
\ell^\star \;=\; \arg\min_{\ell \in \{z_{1},\hat{z}_{1}\}} \;\mathrm{Std}_{b \in [B]}\!\left(\left\|\ell_b\right\|_2\right),
\label{eq:decode_select}
\end{equation}
where $\ell_b$ is the $b$-th sample in the batch $B$, and $\mathrm{Std}(\cdot)$ is the standard deviation over the batch.
Finally, we decode scaled $\ell^\star$ with the frozen VAE decoder:
\[
\tilde X \;=\; D\!\left(\ell^\star / s_{\mathrm{vae}}\right),
\]
where $s_{\mathrm{vae}}$ is the VAE scaling factor.

\section{Experiments}
We now conduct extensive experiments to compare our PGFM algorithm with state-of-the-art methods for dataset distillation.
\subsection{Experimental Setup}
\textbf{Datasets and evaluation.}
To assess the effectiveness of our PGFM approach, we conduct thorough experiments on common dataset distillation benchmarks, specifically focusing on the high-resolution setting of $256 \times 256$ images. The datasets we use are ImageNette and ImageIDC. We utilize the standard hard-label protocol for evaluation, consistent with existing literature on high-resolution dataset distillation. This protocol involves synthesizing the small dataset $\mathcal{S}$ with hard class labels for a specified IPC budget $m \in \{10, 20, 50\}$, then training a target student network (backbone) from scratch on $\mathcal{S}$, and finally evaluating the network's Top-1 accuracy on the original test set of each benchmark. The process is repeated three times to report the mean accuracy, and we apply standard augmentation techniques, such as random resize-crop and CutMix, during the student network's training.

\textbf{Baselines.}
We compare our PGFM method against several state-of-the-art baselines to compare its performance, especially focusing on generative approaches in the latent space. Particularly, we compare against DM \cite{zhao2023dataset} and DiT (the base performance of the pre-trained Diffusion Transformer XL/2 model) \cite{peebles2023scalable}, which showcase the performance of diffusion architectures. Minimax Diffusion (Minimax) \cite{gu2024efficient} is a high-performing generative distillation method that explicitly fine-tunes a pre-trained diffusion model with representative and diversity losses. Our primary state-of-the-art benchmark is MGD$^3$ \citep{chan2025mgd}, a diffusion-based dataset distillation method applying mode-guided sampling to generate a diverse, mode-covering synthetic set. For FM method, we use GMFlow \cite{chen2025gaussian} as our pretrained model, and then directly apply the sampling stage to generate synthetic images. We evaluate performance across three different target student architectures: ConvNet-6, ResNet-18, and ResNet-AP (ResNet with Average Pooling), to ensure a comprehensive assessment of the distilled dataset's generalization ability across model capacities. Results are reported as mean $\pm$ standard deviation over repeated training runs using the same evaluation protocol.

\textbf{Implementation details.}
Our pre-trained model $\mathcal{G}$ is \texttt{gmflow\_imagenet\_k8\_ema} trained on ImageNet, and the image size is 256 x 256.
We use the Flow Matching sampling procedure with a pretrained GMFlow backbone to generate synthetic images. Unless otherwise specified, we run sampling for 48 inference steps with 4 substeps per step, and use probabilistic classifier-free guidance with guidance scale set to $0.3$. We perform prototype discovery using $K$-means++ in latent space with $k=\mathrm{IPC}$, and apply prototype-guided control during sampling with early stopping $s_{\text{end}}=0.6$, path strength $\lambda=0.5$ and a trust-region coefficient $\rho_0=0.5$ with linear decay to $\rho_{min}=0.1$.

\begin{table*}[t]
\centering
\caption{Performance Comparison on ImageNette and ImageIDC Datasets}
\setlength{\tabcolsep}{3.5pt}
\begin{tabular}{l l c c c c c c c c}
\toprule
Dataset & Backbone & IPC & Random & DM & DiT & Minimax & MGD$^3$ & FM & PGFM \\
\midrule
\multirow{9}{*}{Nette}
 & \multirow{3}{*}{ConvNet-6} & $10$ & $42.1 \pm 0.5$ & $48.5 \pm 1.4$ & $55.2 \pm 2.3$ & $56.8 \pm 1.9$ & $58.5 \pm 2.1$ & $60.5 \pm 1.1$ & $\mathbf{63.0 \pm 1.1}$ \\
 &                            & $20$ & $52.6 \pm 0.9$ & $55.2 \pm 0.7$ & $62.3 \pm 1.0$ & $63.9 \pm 0.5$ & $64.3 \pm 1.1$ & $65.3 \pm 2.1$ & $\mathbf{67.6 \pm 1.1}$ \\ 
 &                            & $50$ & $65.2 \pm 1.2$ & $70.4 \pm 0.8$ & $73.8 \pm 0.6$ & $75.4 \pm 0.6$ & $78.1 \pm 2.1$ & $77.7 \pm 1.1$ & $\mathbf{80.4 \pm 1.0}$ \\ 
\cmidrule(lr){2-10}
 & \multirow{3}{*}{ResNet-18} & $10$ & $48.4 \pm 1.0$ & $41.2 \pm 1.0 $ & $59.5 \pm 0.9$ & $61.2 \pm 0.4$ & $62.8 \pm 2.1$ & $61.9 \pm 0.7$ & $\mathbf{64.0 \pm 1.1}$ \\ 
 &                             & $20$ & $58.7 \pm 1.1$ & $51.2 \pm 0.6$ & $63.0 \pm 0.9$ & $65.8 \pm 1.6$ & $68.1 \pm 2.1$ & $67.5 \pm 1.1$ & $\mathbf{70.2 \pm 0.7}$ \\
 &                             & $50$ & $71.3 \pm 2.1$ & $71.2 \pm 0.8$ & $72.7 \pm 0.9$ & $74.5 \pm 0.6$ & $79.1 \pm 1.3$ & $75.1 \pm 1.1$ & $\mathbf{79.2 \pm 1.3}$ \\ 
\cmidrule(lr){2-10}
 & \multirow{3}{*}{ResNet-AP} & $10$ & $48.2 \pm 1.2$ & $40.5 \pm 0.9$ & $56.1 \pm 0.8$ & $58.8 \pm 1.0$ & $65.8 \pm 2.4$ & $62.4 \pm 1.3$ & $\mathbf{66.0 \pm 1.1}$ \\ 
 &                             & $20$ & $55.7 \pm 1.1$ & $52.8 \pm 0.4$ & $63.2 \pm 0.7$ & $65.1 \pm 0.9$ & $70.4 \pm 0.8$ & $65.7 \pm 1.1$ & $\mathbf{70.8 \pm 1.1}$ \\ 
 &                             & $50$ & $68.4 \pm 1.0$ & $70.8 \pm 0.2$ & $72.3 \pm 0.5$ & $74.0 \pm 0.7$ & $\mathbf{79.1 \pm 1.5}$ & $73.6 \pm 0.1$ & $78.8 \pm 1.5$ \\
\midrule
\multirow{9}{*}{IDC}
 & \multirow{3}{*}{ConvNet-6} & $10$ & $41.7 \pm 1.1$ & $46.7 \pm 0.8$ & $47.7 \pm 1.2$ & $47.2 \pm 0.9$ & $49.7 \pm 1.5$ & $51.5 \pm 1.2$ & $\mathbf{53.6 \pm 1.1}$ \\
 &                            & $20$ & $47.0 \pm 1.0$ & $52.6 \pm 0.9$ & $53.2 \pm 1.1$ & $53.2 \pm 0.8$ & $56.0 \pm 1.2$ & $56.7 \pm 1.1$ & $\mathbf{58.8 \pm 1.2}$ \\
 &                            & $50$ & $64.7 \pm 1.3$ & $68.2 \pm 1.0$ & $63.3 \pm 0.9$ & $68.2 \pm 0.7$ & $71.1 \pm 1.5$ & $70.8 \pm 1.3$ & $\mathbf{72.7 \pm 1.2}$ \\
\cmidrule(lr){2-10}
 & \multirow{3}{*}{ResNet-18} & $10$ & $44.9 \pm 1.2$ & $50.2 \pm 0.9$ & $51.3 \pm 1.1$ & $50.7 \pm 0.8$ & $53.4 \pm 1.8$ & $52.6 \pm 0.7$ & $\mathbf{54.3 \pm 1.2}$ \\
 &                            & $20$ & $49.8 \pm 1.1$ & $55.7 \pm 1.0$ & $56.3 \pm 1.2$ & $56.3 \pm 0.7$ & $59.3 \pm 1.5$ & $58.8 \pm 1.4$ & $\mathbf{61.3 \pm 1.0}$ \\
 &                            & $50$ & $65.5 \pm 1.4$ & $69.1 \pm 1.1$ & $64.1 \pm 0.9$ & $69.1 \pm 0.8$ & $72.0 \pm 1.3$ & $68.4 \pm 1.0$ & $\mathbf{72.1 \pm 1.2}$ \\
\cmidrule(lr){2-10}
 & \multirow{3}{*}{ResNet-AP} & $10$ & $47.1\pm1.3$ & $52.8\pm0.5$ & $54.1\pm0.4$ & $53.1\pm0.2$ & $55.9 \pm 2.1$ & $53.2 \pm 0.5$ & $\mathbf{56.3 \pm 0.6}$ \\
 &                             & $20$ & $52.1\pm0.9$ & $58.5\pm0.4$ & $58.9\pm0.2$ & $59.0\pm0.4$ & $61.9 \pm 0.9$ & $57.4 \pm 1.3$ & $\mathbf{62.1 \pm 1.2}$ \\
 &                             & $50$ & $66.1\pm1.8$ & $69.1\pm0.8$ & $64.3\pm0.6$ & $69.6\pm0.2$ & $\mathbf{72.1 \pm 0.8}$ & $66.8 \pm 0.9$ & $71.7 \pm 1.0$ \\
\bottomrule
\end{tabular}
\label{tab:main}
\end{table*}

\subsection{Result Analysis}


\begin{table*}[t]
\centering
\small
\setlength{\tabcolsep}{7pt}
\caption{Performance Comparison on ImageNet100.}
\label{tab:imagenet100}
\begin{tabular}{l|ccc|ccc}
\toprule
\multirow{2}{*}{} & \multicolumn{3}{c|}{$\text{IPC}=10$ (0.8\%)} & \multicolumn{3}{c}{$\text{IPC}=20$ (1.6\%)} \\
\midrule
 & ConvNet-6 & ResNet-18 & ResNetAP-10 & ConvNet-6 & ResNet-18 & ResNetAP-10 \\
\midrule
Random & $17.0\pm0.3$ & $19.1\pm0.4$ & $17.5\pm0.5$ & $24.8\pm0.2$ & $26.7\pm0.5$ & $25.5\pm0.3$ \\
Herding  & $17.2\pm0.3$ & $19.8\pm0.3$ & $16.1\pm0.2$ & $24.3\pm0.4$ & $27.6\pm0.1$ & $24.7\pm0.1$ \\
IDC-1  & $24.3\pm0.5$ & $25.7\pm0.1$ & $25.1\pm0.2$ & $28.8\pm0.3$ & $29.9\pm0.2$ & $30.2\pm0.2$ \\
MinMaxDiff  & $22.3\pm0.5$ & $24.8\pm0.2$ & $22.5\pm0.3$ & $29.3\pm0.4$ & $32.3\pm0.1$ & $31.2\pm0.1$ \\
MGD$^{3}$ (Ours) & $23.4\pm0.9$ & $23.8\pm0.5$ & $25.7\pm0.7$ & $30.2\pm0.5$ & $32.0\pm1.4$ & $33.1\pm0.5$ \\
\midrule
FM & $24.9\pm0.4$ & $24.8\pm0.7$ & $25.5\pm0.4$ & $30.6\pm0.2$ & $32.0\pm0.4$ & $31.5\pm0.6$ \\
PGFM (Ours) & $\mathbf{25.7}\pm0.4$ & $\mathbf{26.5}\pm0.4$ & $\mathbf{27.2}\pm0.3$ & $\mathbf{31.2}\pm0.2$ & $\mathbf{32.9}\pm0.5$ & $\mathbf{34.0}\pm0.5$ \\
\midrule
Full & $79.9\pm0.4$ & $80.3\pm0.2$ & $81.8\pm0.7$ & $79.9\pm0.4$ & $80.3\pm0.2$ & $81.8\pm0.7$ \\
\bottomrule
\end{tabular}
\end{table*}

As shown in Table~\ref{tab:main}, PGFM almost always outperforms prior diffusion-based distillation baselines on both ImageNette and ImageIDC across backbones, with the largest gains under tight budgets, i.e., low IPC. On ImageNette, PGFM improves over the strongest diffusion baseline MGD$^3$ most at $\text{IPC}=10/20$ (e.g., by about $+4.5$ points on ConvNet-6 at $\text{IPC}=10$ and  $+2.1$ points on ResNet18 at $\text{IPC}=20$), and remains competitive at $\text{IPC}=50$ where the synthesis budget is less constrained. 
On ImageIDC, the same pattern holds and further highlights robustness: FM sampling is already strong, and PGFM provides additional gains at $\text{IPC}=10/20$ (by roughly $+1$--$2$ points over MGD$^3$ in several cases) while remaining competitive at $\text{IPC}=50$. 
This trend underscores the core advantage of PGFM: by explicitly guiding flow trajectories toward diverse prototypes, our method ensures that even a minimal set of synthetic samples can effectively cover the critical modes of the data distribution.

\paragraph{ImageNet100.}
Table~\ref{tab:imagenet100} reports results on the more challenging ImageNet100 benchmark. When $\text{IPC}=10$, pure flow-matching sampling (FM, no guidance) already improves over strong diffusion baselines (e.g., beating MGD$^{3}$ by $+1.5$ and $+1.0$ points on ConvNet-6 and ResNet-18, respectively), suggesting that flow-based sampling produces latents that are inherently closer to the class-conditional manifold and thus more “trainable” under scarce budgets. With this strong starting point, PGFM further boosts FM by $+0.8$, $+1.7$, and $+1.7$ points across the three backbones, respectively, validating our core idea that a lightweight, path-controlled guidance improves mode coverage while avoiding late-stage detail degradation. At $\text{IPC}=20$, the gaps naturally narrow as coverage improves with more samples, but PGFM remains consistently better than both FM and MGD$^{3}$.

\begin{table}[t]
\centering
\caption{ImageNet-1K image-to-image accuracy (\%) under $\text{IPC}=10$.}
\setlength{\tabcolsep}{3.5pt}
\renewcommand{\arraystretch}{1.2}
\begin{tabular}{lccc}
\hline
Method & ConvNet-6 & ResNet18 & ResNet-AP \\
\hline
MiniMaxDiffusion  & 11.6 & 17.7 & 16.6 \\
FM          & 12.9 & 19.8 & 19.4 \\
MGD3              & 13.6 & 21.5 & 20.2 \\
PGFM              & \textbf{14.1} & \textbf{22.8} & \textbf{21.3} \\
\hline
\end{tabular}
\vspace{0.3em}
\label{tab:imagenet1k}
\end{table}

\paragraph{ImageNet1k.}
Table~\ref{tab:imagenet1k} summarizes ImageNet-1K image-to-image results under the extremely tight budget of $\text{IPC}=10$. Even without diffusion-style sampling, pure flow-matching (FM) consistently outperforms MiniMaxDiffusion across all backbones, with gains of $+1.3$, $+2.1$, and $+2.8$ points, respectively. MGD3 improves over FM on ResNet18 but remains close on ConvNet-6 and ResNet-AP. Our method, PGFM, achieves the best performance on all three architectures, which corresponds to improvements over MGD3 of $+0.5$, $+2.7$, and $+1.3$ points, respectively. These results indicate that path-guided flow sampling is particularly effective in the low-IPC regime on large-scale datasets.

\paragraph{Computational cost.}

\begin{table}[t]
\centering
\caption{Computational Complexity Comparison, with metrics M: Millions, NFE: Number of Function Evaluations, T: Trillions. $\downarrow$ means less is better.}
\resizebox{0.9\linewidth}{!}{%
\begin{tabular}{lccc}
\toprule
Metric & MiniMaxDiff & MGD$^3$ & PGFM\\
\midrule
Params (M) $\downarrow$      & 674.93 & 675.24 & \textbf{542.77} \\
Steps (NFE) $\downarrow$     & 50     & 50     & \textbf{32}     \\
FLOPs (T) $\downarrow$       & 23.73  & 17.30  & \textbf{2.27}   \\
\midrule
\textit{Efficiency Gain} & 1.0$\times$ & 1.4$\times$ & \textbf{10.5$\times$} \\
\bottomrule
\end{tabular}%
}
\label{tab:complexity}
\end{table}

\vspace{-0.7em}

Table~\ref{tab:complexity} compares the computational costs of PGFM against state-of-the-art diffusion-based distillation methods. PGFM demonstrates superior efficiency across all metrics. Specifically, our model requires significantly fewer parameters ($542.77$M) compared to the heavier DiT-based baselines ($\sim 675$M). Notably, PGFM drastically reduces the computational burden, operating at only $2.27$ TFLOPs per image—an order of magnitude lower than MiniMaxDiff ($23.73$ TFLOPs) and MGD$^3$ ($17.30$ TFLOPs). This results in a substantial $10.5\times$ efficiency gain relative to MiniMaxDiff, validating that PGFM is not only effective but also computationally lightweight.

\paragraph{Complex models.}

\begin{table}[t]
\centering
\caption{Accuracy (\%) on ImageNet-1K (ResNet-18/50/101), $\text{IPC}=10$.}
\label{tab:deeper}
\setlength{\tabcolsep}{3.5pt}
\begin{tabular}{lccc}
\toprule
Method & ResNet-18 & ResNet-50 & ResNet-101 \\
\midrule
MiniMaxDiffusion & 17.7 & 19.2 & 18.4 \\
MGD3           & 20.5 & 22.5 & 22.1   \\
FM             & 19.8 & 21.4 & 20.5 \\
PGFM             & \textbf{22.8} & \textbf{23.8} & \textbf{22.3} \\
\bottomrule
\end{tabular}
\end{table}

Table~\ref{tab:deeper} shows that under $\text{IPC}=10$ on ImageNet-1K, PGFM achieves the best accuracy across ResNet-18/50/101, surpassing MGD3 by a clear margin. Moreover, the unguided FM baseline is already competitive with diffusion baselines (e.g., on ResNet-50/101), suggesting that flow-matching sampling produces more transferable and trainable synthetic data under tight budgets, and PGFM further improves with lightweight prototype-guided coverage.

\subsection{Ablation Study}

\paragraph{Component analysis.}
\begin{table}[t]
\centering
\small
\setlength{\tabcolsep}{2.5pt}
\caption{Ablation study on ImageNette, $\text{IPC}=10$. Path C means path control. Early S means early stop. Trust R means trust region.}
\label{tab:ablation}
\begin{tabular}{l c c c c c}
\toprule
\textbf{Model} & \textbf{Path C.} & \textbf{Early S.} & \textbf{Trust R.} & \textbf{Acc.} \\
\midrule
ConvNet-6     & \multirow{3}{*}{--} & \multirow{3}{*}{--}             & \multirow{3}{*}{--}             & $60.5\pm2.1$ \\
ResNetAP-10   &                               &                                 &                                 & $61.9\pm1.7$ \\
ResNet-18     &                               &                                 &                                 & $62.4\pm2.3$ \\
\midrule
ConvNet-6     & \multirow{3}{*}{$\checkmark$} & \multirow{3}{*}{--}             & \multirow{3}{*}{--}             & $61.6\pm1.2$ \\
ResNetAP-10   &                               &                                 &                                 & $62.4\pm1.7$ \\
ResNet-18     &                               &                                 &                                 & $63.0\pm1.5$ \\
\midrule
ConvNet-6     & \multirow{3}{*}{$\checkmark$} & \multirow{3}{*}{$\checkmark$}   & \multirow{3}{*}{--}             & $62.1\pm1.3$ \\
ResNetAP-10   &                               &                                 &                                 & $63.5\pm1.6$ \\
ResNet-18     &                               &                                 &                                 & $64.8\pm2.0$ \\
\midrule
ConvNet-6     & \multirow{3}{*}{$\checkmark$} & \multirow{3}{*}{$\checkmark$}   & \multirow{3}{*}{$\checkmark$}   & $\mathbf{63.0}\pm1.1$ \\
ResNetAP-10   &                               &                                 &                                 & $\mathbf{64.0}\pm1.1$ \\
ResNet-18     &                               &                                 &                                 & $\mathbf{66.0}\pm1.1$ \\
\bottomrule
\end{tabular}
\end{table}

Table~\ref{tab:ablation} reports an ablation on ImageNette with $\text{IPC}=10$ across three evaluation backbones. Starting from unguided FM sampling, adding path control alone already yields consistent gains (up to $+1.1$ points on ResNet-18). However, path guidance can partially disrupt the underlying flow field, sometimes introducing blur or noise and leading to higher variance in downstream task. To mitigate it, we introduce early stop, which turns off guidance in the later phase so that the base flow can recover fine details. This further improves performance, especially on ResNet-18. Finally, the trust region for guidance stabilizes sampling by bounding the control strength relative to the model dynamics, preventing over-steering into poorly supported regions of the latent space. Combining all three components gives the most stable results, reaching the best performance. 
\paragraph{Coverage analysis.}

\definecolor{darkblue}{RGB}{31, 119, 180}   
\definecolor{lightblue}{RGB}{198, 219, 239} 
\definecolor{graycross}{RGB}{110, 110, 110} 

\newcommand{\legenddot}[1]{%
    \tikz[baseline=-0.6ex]{
        \fill[#1] (0,0) circle (3.5pt); 
    }%
}

\newcommand{\legendcross}[1]{%
    \tikz[baseline=-0.6ex]{
        \draw[#1, line width=1.5pt] (-3.5pt,-3.5pt) -- (3.5pt,3.5pt);
        \draw[#1, line width=1.5pt] (-3.5pt,3.5pt) -- (3.5pt,-3.5pt);
    }%
}

\begin{figure}[t]
    \centering
    \begin{subfigure}[b]{0.49\linewidth}
        \centering
        \includegraphics[width=\linewidth]{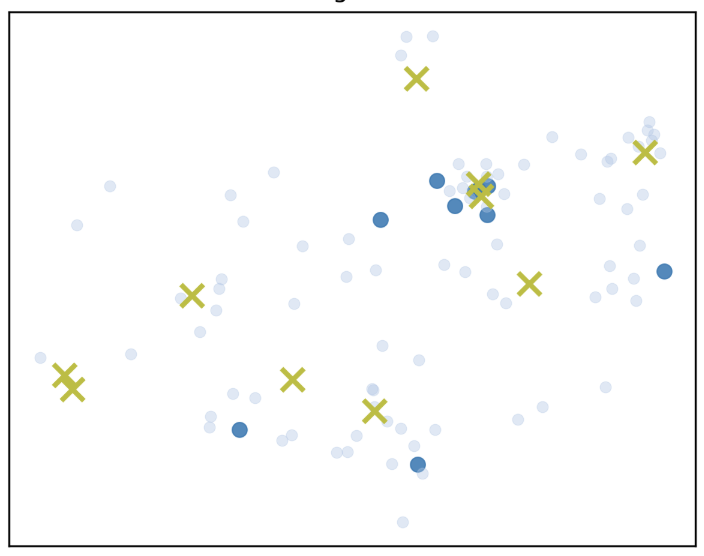}
        \caption{Baseline FM}
        \label{fig:baseline}
    \end{subfigure}
    \hfill
    \begin{subfigure}[b]{0.49\linewidth}
        \centering
        \includegraphics[width=\linewidth]{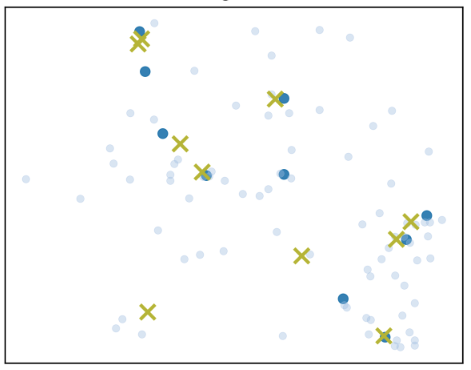}
        \caption{MGD$^3$}
        \label{fig:mgd3}
    \end{subfigure}
    
    \vspace{1em} 

    \begin{subfigure}[b]{0.49\linewidth}
        \centering
        \includegraphics[width=\linewidth]{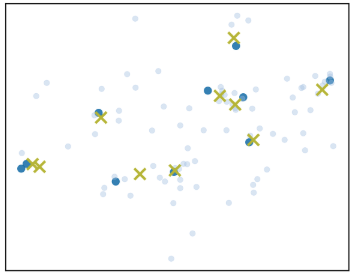}
        \caption{PGFM (Ours)}
        \label{fig:pgfm}
    \end{subfigure}
    \hfill
    \begin{subfigure}[b]{0.495\linewidth}
        \centering
        \includegraphics[width=\linewidth]{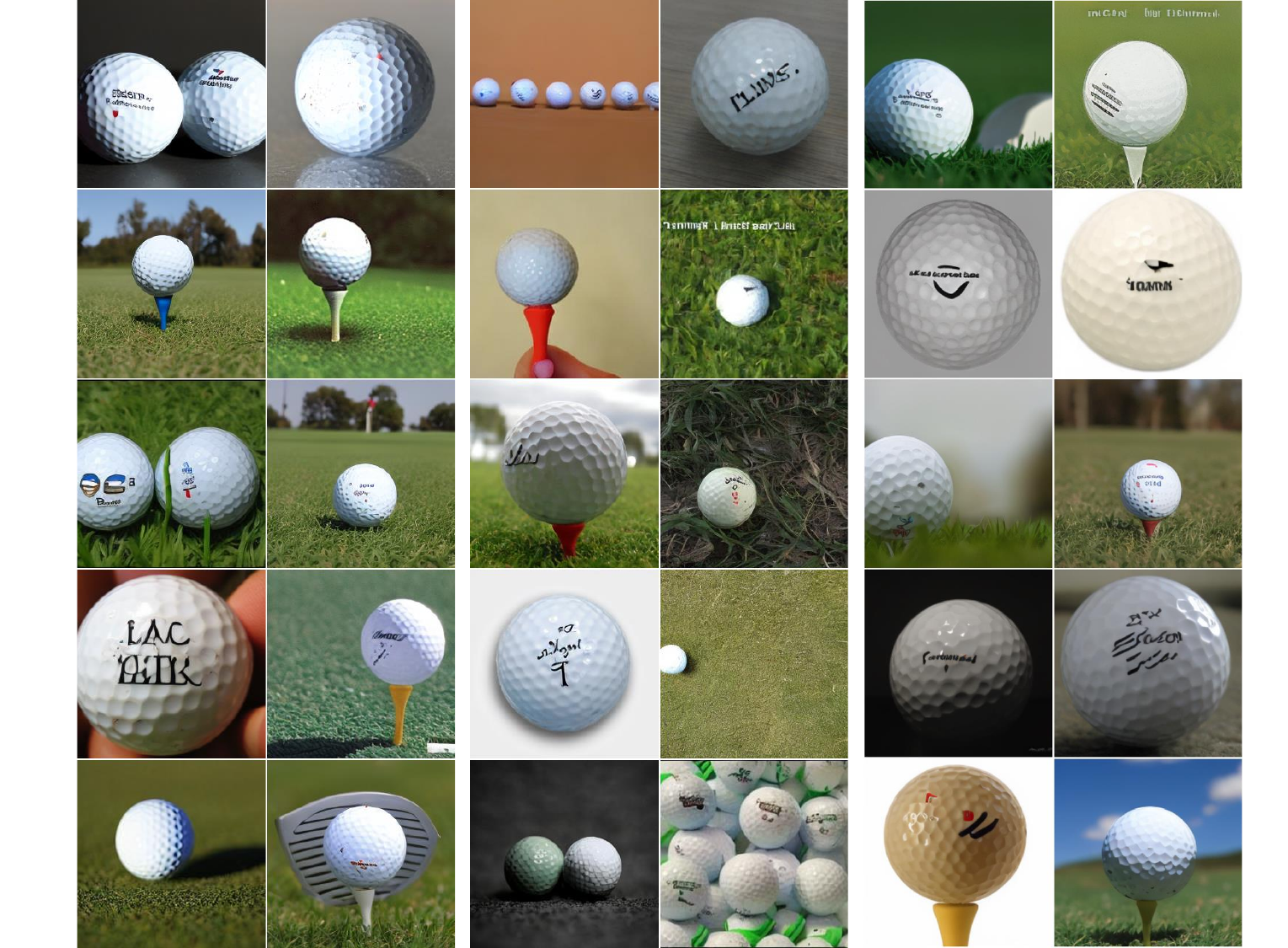}
        \caption{FM, MGD$^3$, PGFM}
        \label{fig:pgfm}
    \end{subfigure}

    \caption{\textbf{t-SNE Analysis.} 
    Visual comparison of latent distributions.
    (a) Generated by FM baseline. 
    (b) Generated by MGD$^3$. 
    (c) Generated by our PGFM.
    (\legenddot{darkblue} Synthetic image, \legenddot{lightblue} Real image, \legendcross{graycross} Prototypes). 
    The PGFM method demonstrates better alignment with the real latent distribution.}
    \label{fig:tSNE}
\end{figure}

To illustrate diversity coverage in latent space, we visualize the distilled datasets with t-SNE for MGD$^3$, FM, and PGFM. As shown in Figure~\ref{fig:tSNE}, the unguided FM baseline exhibits noticeably weaker coverage, with synthetic samples occupying only part of the real-data manifold. MGD$^3$ improves coverage by applying mode guidance, but some synthetic features still drift away from their intended prototypes, indicating imperfect prototype-to-sample alignment. In contrast, PGFM yields synthetic embeddings that cluster around the assigned prototypes and overlap more consistently with the real-data distribution. This suggests that our path-guided control provides a more reliable mechanism for mode assignment and coverage: by encouraging the sampler to visit diverse class-conditional regions early in the trajectory, it yields a diverse distilled dataset which is better aligned with the underlying data manifold.

\begin{table}[t]
\centering
\caption{Coverage analysis on ImageNette, $\text{IPC}=10$. Warm S means warm start. Hit rate (\%) measures prototype coverage, which is the percentage of samples whose nearest prototype equals the assigned one.}
\resizebox{\linewidth}{!}{%
\begin{tabular}{lcc|cccc}
\toprule
Component & FM & MGD$^3$ & \multicolumn{4}{c}{PGFM} \\
\midrule
Path C.  & -- & -- & \checkmark & \checkmark & \checkmark & \checkmark \\
Early S. & -- & -- & --         & \checkmark & \checkmark & \checkmark \\
Trust R. & -- & -- & --         & --         & \checkmark & \checkmark \\
Warm S.  & -- & -- & --         & --         & --         & \checkmark \\
\midrule
\textit{Coverage/Hit rate (\%)} & 10 & 30 & 18 & 27 & 33 & \textbf{78} \\
\bottomrule
\end{tabular}%
}
\label{tab:coverage}
\end{table}

\vspace{-0.5em}

Table~\ref{tab:coverage} shows that unguided FM has very poor prototype coverage ($10\%$), while MGD$^3$ achieves $30\%$ but still mismatches many prototypes. For PGFM, path control alone helps but can perturb the original flow field and drift off-manifold. Although adding the early stop and trust region stabilizes the guidance, coverage remains limited. Warm start is the key to entering the correct prototype basin early, giving rise to a large jump in hit rate (78\%) and enabling the flow to match most cluster centers reliably (see Appendix \ref{append:ablation} for more details).

\paragraph{Representativeness and coverage.}

\begin{figure}[t]
    \centering
    \includegraphics[width=\linewidth]{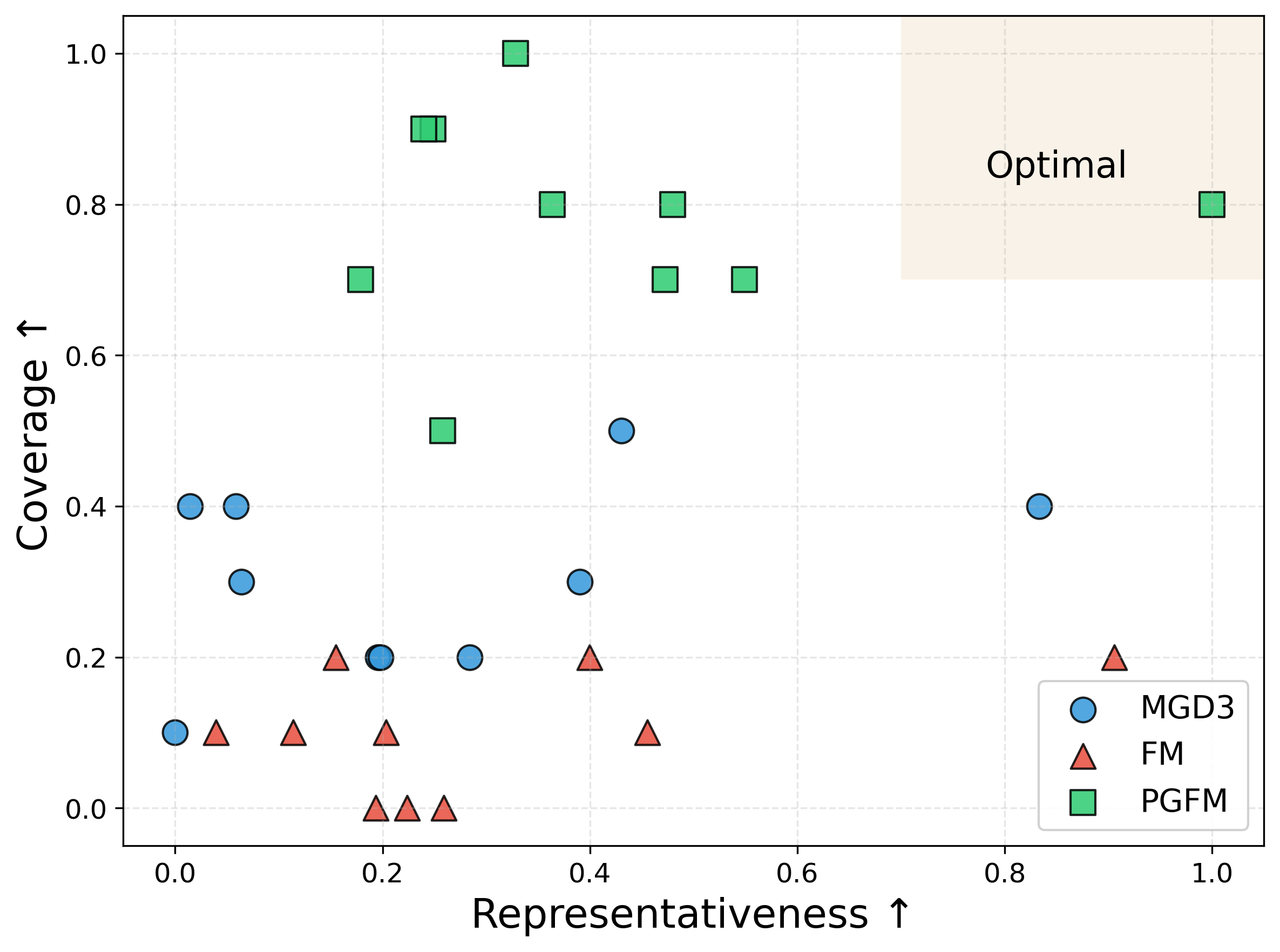}
    \caption{Representativeness and coverage analysis
    }
    \label{fig:rep_cov}
\end{figure}

\vspace{-0.4em}

We visualize the trade-off between representativeness and coverage in Figure~\ref{fig:rep_cov}: representativeness is measured by calculating the average distance from the synthetic sample to its $50$ nearest neighbors in the real dataset and converting it into a score, while coverage quantifies the diversity. Higher values indicate better performance for both measurements.
The region in the top-right corner denotes the Optimal balance. We can see that PGFM achieves significantly superior coverage, consistently populating the upper region of the plot. Meanwhile, PGFM demonstrates improved representativeness, with several class prototypes successfully landing in or near the optimal region, outperforming MGD$^3$ and FM (see Appendix \ref{append:exp} for more experiments).

\section{Conclusion}
In this paper, we present PGFM, a simple and efficient flow-matching framework for generative dataset distillation. By operating in a compressed VAE latent space and sampling from a pretrained flow-based generator via ODE integration, PGFM avoids the heavy compute overhead of diffusion-based synthesis. We further introduce lightweight path-controlled guidance applied only in the early sampling stage, improving mode coverage without washing out fine-grained details. As a result, PGFM can synthesize datasets with only a small number of solver steps, while maintaining strong fidelity and diversity. Experiments on challenging benchmarks such as ImageNette and ImageIDC demonstrate that PGFM is competitive with, and often improves over, strong diffusion-based baselines, highlighting flow matching as a practical alternative for scalable dataset distillation.

\newpage

\bibliographystyle{icml2026} 
\bibliography{example_paper} 

\appendix
\onecolumn

\section{Hyperparameter Setting}
\label{append:hyper}
 
We set the classifier-free guidance scale to $0.3$. For the flow matching ODE solver, $N=48$ inference steps with $4$ intermediate sub-steps are used to ensure high trajectory precision while maintaining computational efficiency. The latent prototypes are initialized using $K$-means++ clustering by default, where we employ cluster centroids as guidance targets with convergence tolerance 1e-4 and up to 300 iterations. To facilitate convergence, particularly for ResNet-based evaluations, we introduce an initialization bias $\eta_{init}$ that pushes the starting noise $x_T$ towards the assigned prototype. The value of $\eta_{init}$ is adaptively adjusted based on dataset complexity and IPC. Specifically, for ImageNette, we set $\eta_{init} \in \{0.09, 0.05, 0.01\}$ for IPCs $\{10, 20, 50\}$, respectively. For the more challenging ImageIDC and ImageNet-100 datasets, we employ smaller bias values of $\{0.04, 0.02, 0.01\}$ for the corresponding IPC settings. Finally, the scope of the guidance force is controlled by a terminal trust region radius $\rho_{min}=0.1$, which determines the linear decay schedule in conjunction with the initial radius $\rho_0$.

\section{Extensive Experiments}
\label{append:exp}

\subsection{Effect of $\eta_{init}$}
\label{append:ablation}

Table~\ref{tab:eta_ablation_hitrate} shows that the initialization strength $\eta_{\text{init}}$ has a strong impact on prototype coverage. As $\eta_{\text{init}}$ increases, the Hit Rate rises steadily (from 33\% at $\eta_{\text{init}}{=}0$ to 86\% at $\eta_{\text{init}}{=}0.15$), indicating that warm-starting closer to the assigned prototype helps the trajectory enter the correct mode basin early. However, larger $\eta_{\text{init}}$ is not always better: overly aggressive initialization can over-constrain samples toward the prototype centers, reducing discriminative details and thereby hurting downstream task. As a result, $\eta_{\text{init}}$ controls a clear trade-off between coverage and sample utility, and we find a moderate value ($\eta_{\text{init}}{=}0.09$) which achieves the best overall accuracy while substantially improving coverage on ImageNette for $\text{IPC}=10$. Additional hyperparameter settings are provided in Appendix~\ref{append:hyper}.

\begin{table}[h]
\centering
\caption{Ablation study on initialization parameter $\eta_{init}$. We report both accuracy and Hit Rate (coverage) on ImageNette ($\text{IPC}=10$).}
\label{tab:eta_ablation_hitrate}
\setlength{\tabcolsep}{5pt} 
\begin{tabular}{lcccc}
\toprule
\multirow{2}{*}{\textbf{$\eta_{init}$}} & \multicolumn{3}{c}{Accuracy (\%)} & \multirow{2}{*}{Hit Rate(\%)} \\
\cmidrule(lr){2-4} 
 & ConvNet & ResNet18 & ResNet-AP &  \\
\midrule
0     & 62.1 $\pm$ 1.3 & 63.5 $\pm$ 1.6 & 64.8 $\pm$ 2.0 & 33 \\
0.05  & 61.6 $\pm$ 1.3 & 62.3 $\pm$ 0.9 & 63.8 $\pm$ 1.2 & 52 \\
0.09  & \textbf{63.0 $\pm$ 1.1} & \textbf{64.0 $\pm$ 1.1} & \textbf{66.0 $\pm$ 1.1} & 78 \\
0.15  & 62.0 $\pm$ 1.3 & 62.8 $\pm$ 0.8 & 64.2 $\pm$ 1.2 & \textbf{86} \\
\bottomrule
\end{tabular}
\end{table}

\subsection{Details of Hit Rate}
\label{append:coverage}

Table~\ref{tab:per_class_hit_rate} shows the detailed hit rate of each class when comparing with FM and MGD$^3$. Our method, PGFM, can significantly increase the coverage, demonstrating that our method is better aligned with the prototypes to achieve the diversity.

\begin{table}[t]
\centering
\caption{Per-class Hit Rate (\%) comparison on ImageNette ($\text{IPC}=10$). PGFM achieves significantly higher coverage across all distinct semantic classes.}
\begin{tabular}{llccc}
\toprule
\textbf{Class ID} & \textbf{Class Name} & \textbf{FM} & \textbf{MGD$^3$} & \textbf{PGFM (Ours)} \\
\midrule
n01440764 & Tench             & 10.0 & 10.0 & \textbf{80.0} \\
n02102040 & English Springer  & 20.0 & 30.0 & \textbf{70.0} \\
n02979186 & Cassette Player   & 0.0  & 20.0 & \textbf{50.0} \\
n03000684 & Chainsaw          & 10.0 & 30.0 & \textbf{70.0} \\
n03028079 & Church            & 20.0 & 20.0 & \textbf{80.0} \\
n03394916 & French Horn       & 10.0 & 40.0 & \textbf{90.0} \\
n03417042 & Garbage Truck     & 0.0  & 40.0 & \textbf{70.0} \\
n03425413 & Gas Pump          & 10.0 & 10.0 & \textbf{90.0} \\
n03445777 & Golf Ball         & 0.0  & 60.0 & \textbf{100.0} \\
n03888257 & Parachute         & 20.0 & 40.0 & \textbf{80.0} \\
\midrule
\textbf{Average} & --           & \textbf{10.0} & \textbf{30.0} & \textbf{78.0} \\
\bottomrule
\end{tabular}%
\label{tab:per_class_hit_rate}
\end{table}

\subsection{Cluster Closest Point}

Table~\ref{tab:closest_point} shows that using closest point instead of the cluster center will significantly reduce the performance. This is because it turns a stable mode representative (the centroid) into a single, noisy sample. The centroid summarizes the cluster’s high-density region and provides a smooth, robust target for guidance, whereas the closest point still carries instance-level details and sample noise. Guiding trajectories toward such samples tends to overfit specific appearances and reduces within-class diversity.

\begin{table}[h]
\centering
\caption{Ablation study on using closest point instead of centroid, with ImageNette, $\text{IPC}=10$.}
\label{tab:closest_point}
\setlength{\tabcolsep}{8pt}
\begin{tabular}{lccc}
\toprule
\textbf{Method} & \textbf{ConvNet} & \textbf{ResNet18} & \textbf{ResNet-AP} \\
\midrule
Closest Point & 61.1 $\pm$ 1.2 & 61.3 $\pm$ 1.4 & 62.8 $\pm$ 0.9 \\
PGFM (Ours)   & \textbf{63.0 $\pm$ 1.1} & \textbf{64.0 $\pm$ 1.1} & \textbf{66.0 $\pm$ 0.1} \\
\bottomrule
\end{tabular}
\end{table}

\subsection{Other Cluster Algorithm}

We further investigate the impact of different clustering algorithms by comparing our approach against K-means\cite{likas2003global}, Agglomerative Clustering \cite{ackermann2014analysis} and Gaussian Mixture Models (GMM) \cite{rasmussen1999infinite}. The results in Table~\ref{tab:clustering_ablation} demonstrate that the $K$-means++ clustering algorithm consistently yields the best performance.

\begin{table}[t]
\centering
\caption{Ablation study on different clustering algorithms, with ImageNette, $\text{IPC}=10$}
\label{tab:clustering_ablation}
\setlength{\tabcolsep}{8pt} 
\begin{tabular}{lccc}
\toprule
\textbf{Method} & \textbf{ConvNet} & \textbf{ResNet18} & \textbf{ResNet-AP} \\
\midrule
GMM & $61.6 \pm 1.2$ & $64.0 \pm 0.9$ & $63.6 \pm 1.1$ \\
Agglomerative & $61.4 \pm 1.3$ & $63.4 \pm 1.0$ & $64.0 \pm 1.2$ \\
K-means & $62.5 \pm 1.1$ & $63.8 \pm 1.1$ & $64.8 \pm 0.1$ \\
K-means++ & $\mathbf{63.0 \pm 1.1}$ & $\mathbf{64.0 \pm 1.1}$ & $\mathbf{66.0 \pm 0.1}$ \\
\bottomrule
\end{tabular}
\end{table}

\section{Algorithm}
\label{append:alg}

The pseudo code of the PGFM algorithm is given in Algorithm \ref{alg:pgfm}.

\begin{algorithm}[tb]
\caption{PGFM: Path-Guided Flow Matching Sampling}
\label{alg:pgfm}
\begin{algorithmic}[1]
\REQUIRE Budget IPC; frozen VAE Decoder $D$; pretrained GMFlow velocity field $u_\phi$; class-wise prototypes $\{\mu_{y,k}\}$; guidance strength $\lambda$; early-stop $g$; trust-region $\rho_0$; init pull $\eta_{\mathrm{init}}$.
\ENSURE Distilled synthetic set $\tilde{\mathcal{S}}$ with IPC samples per class.

\vspace{0.2em}
\STATE \textbf{Flow-matching sampling by GMFlow with path guidance }
\STATE $\tilde{\mathcal{S}} \leftarrow \emptyset$
\FOR{each class $y$}
    \FOR{$k=1$ to $\mathrm{IPC}$}
        \STATE Sample base latent $z \sim \mathcal{N}(0,I)$

        \IF{$\eta_{\mathrm{init}} > 0$}
            \STATE Warm-start before guidance using Eq.~\eqref{eq:warm_start}
        \ENDIF

        \FOR{$t\in [0,1]$}
            \STATE Evaluate base velocity using Eq.~\eqref{eq:ode}
            \STATE Set early-stop gate $g$ with Eq.~\eqref{eq:early_stop}
            \IF{$g=1$}
                \STATE Compute path control velocity with Eq.~\eqref{eq:proto_control}
                \STATE Apply trust-region scale $\alpha$ with Eq.~\eqref{eq:trust_region}
                \STATE Get guided velocity $\tilde u \leftarrow u_\phi + \alpha\,u_{\mathrm{proto}}$
            \ELSE
                \STATE Get guided velocity $\tilde u \leftarrow u_\phi$
            \ENDIF
            \STATE Integrate the guided ODE form to get the synthetic features
        \ENDFOR
        
        \STATE Decode synthetic image $\tilde x \leftarrow D(z)$
        \STATE $\tilde{\mathcal{S}} \leftarrow \tilde{\mathcal{S}} \cup \{(\tilde x,y)\}$
    \ENDFOR
\ENDFOR
\STATE \textbf{return} $\tilde{\mathcal{S}}$
\end{algorithmic}
\end{algorithm}

\section{Visualization}

\subsection{Effect of Trust Region $\rho$}
\label{append:bulr}

\begin{figure}[t]
    \centering
    \includegraphics[width=0.6\linewidth]{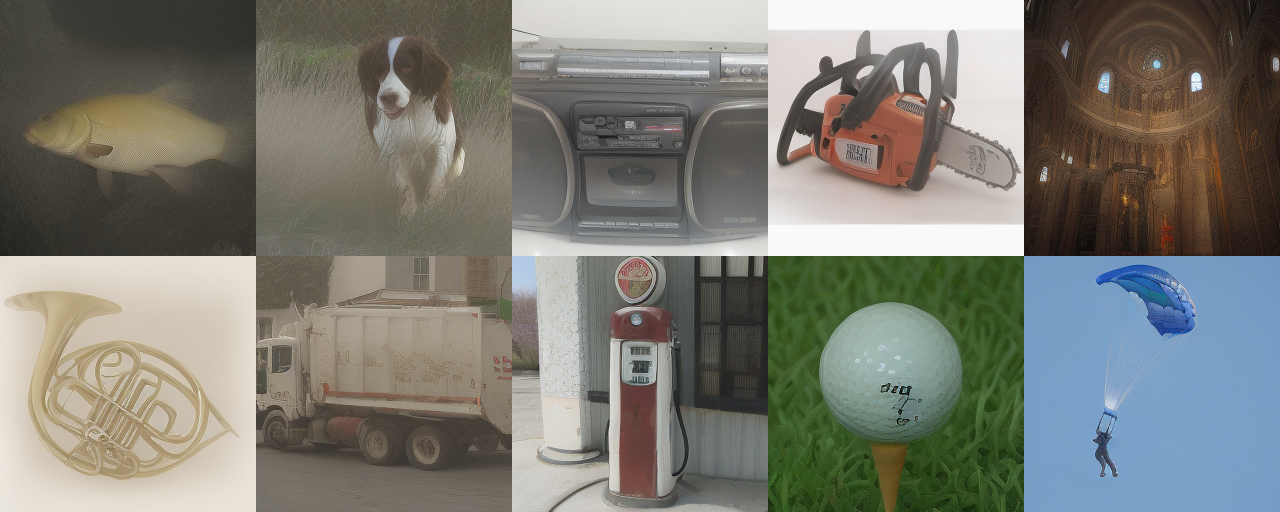}
    \caption{
    Blurring synthetic images on ImageNette for $\text{IPC}=10$    
    }
    \label{fig:blur}
\end{figure}

Figure~\ref{fig:blur} shows that when we set a high $\rho_0$ or use a constant $\rho$ throughout sampling, the generated images become noticeably blurred. This indicates that overly heavy path-guided control can dominate the learned flow field and steer the trajectory away from the model’s natural transport manifold, where the vector field and the frozen VAE decoder are less reliable.

\subsection{Synthetic Images}

\begin{figure}[t]
    \centering
    \includegraphics[width=\linewidth]{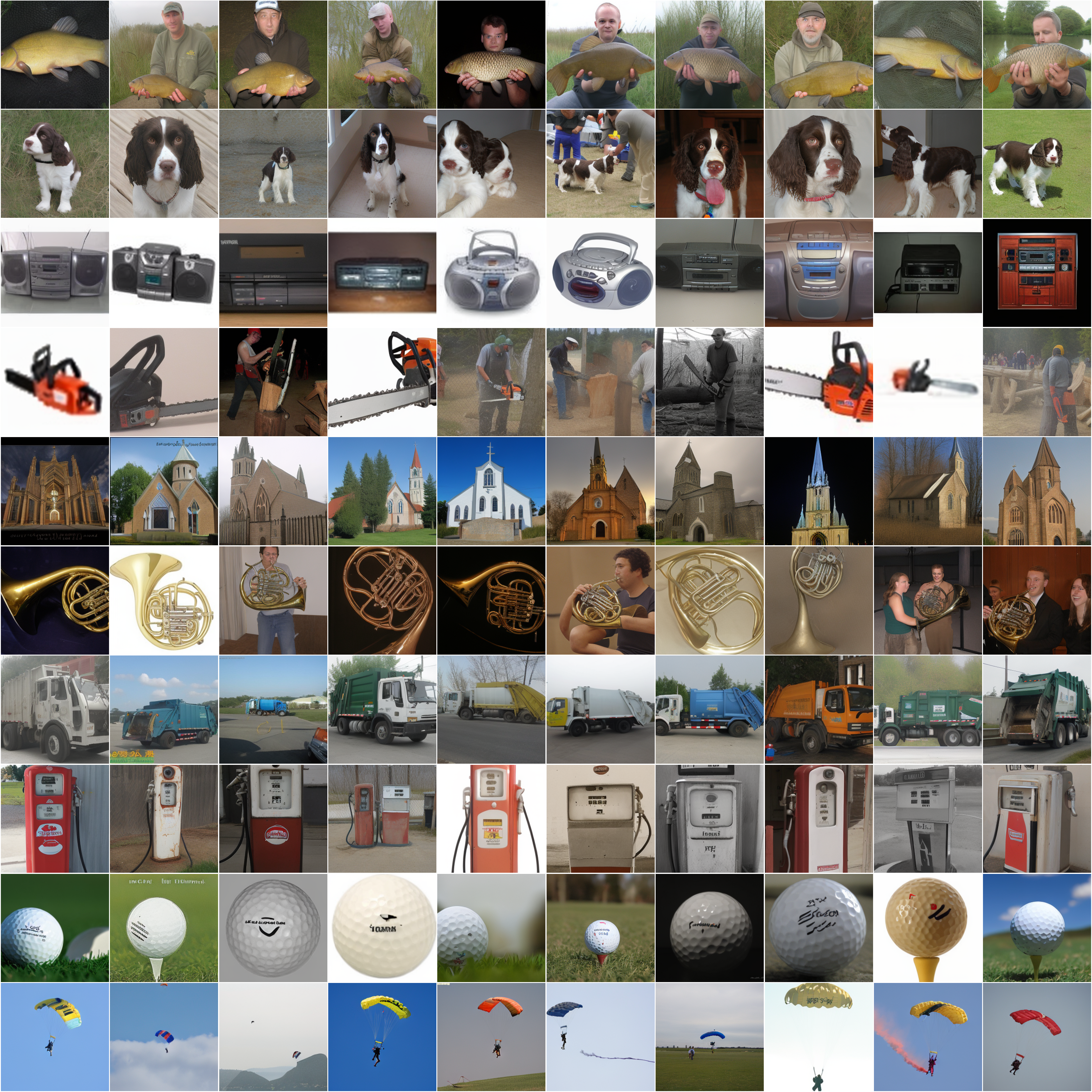}
    \caption{
    Synthetic images with ImageNette and $\text{IPC}=10$    
    }
    \label{fig:allimages}
\end{figure}

Figure~\ref{fig:allimages} shows the synthetic images that are generated on ImageNette for $\text{IPC}=10$.

\subsection{Flow Matching versus Diffusion}

\begin{figure}[t]
    \centering
    \includegraphics[width=0.8\linewidth]{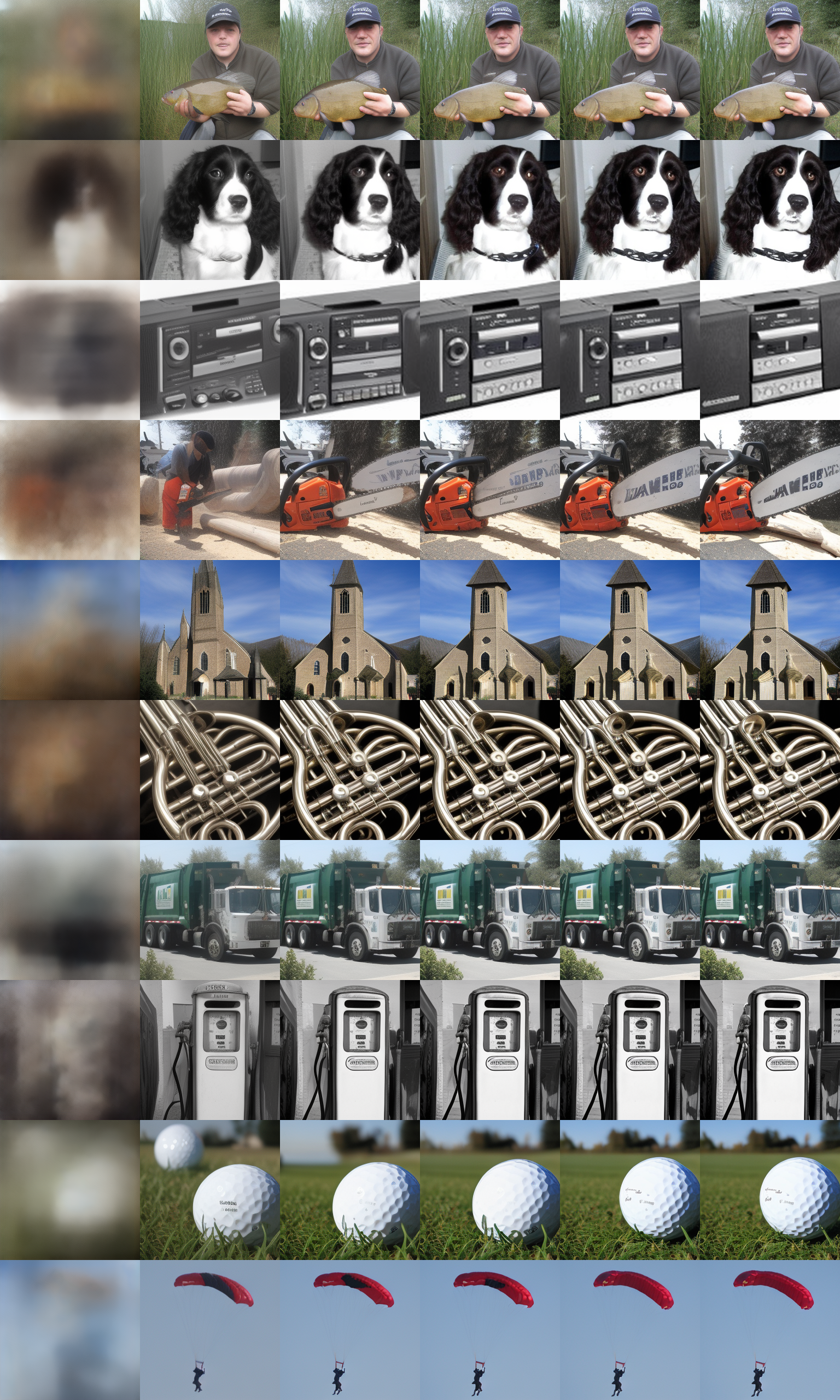}
    \caption{
    Visualization of the flow matching-based sampling step from time 0 to 1.
    }
    \label{fig:flowimages}
\end{figure}

\begin{figure}[t]
    \centering
    \includegraphics[width=0.8\linewidth]{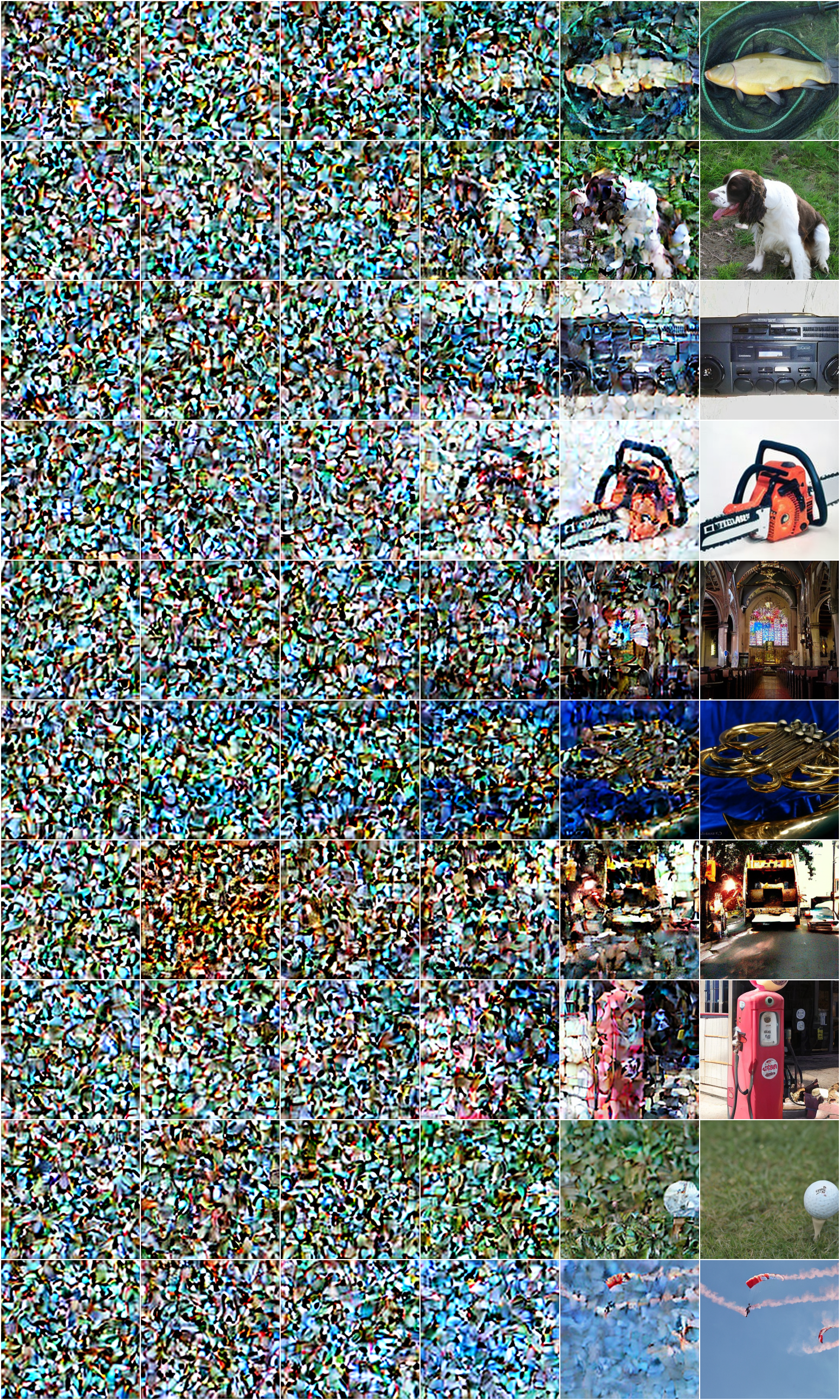}
    \caption{
    Visualization of the diffusion-based sampling.
    }
    \label{fig:diffimages}
\end{figure}

A comparison of the generation trajectories in Figure~\ref{fig:flowimages} highlights the superior efficiency of flow matching. We can see that the flow matching process yields a smooth evolution where recognizable structure emerges early in the trajectory, even under path-guided control. In contrast, the diffusion process in Figure~\ref{fig:diffimages} remains dominated by noise for the majority of the sampling timeline, with meaningful semantic content only materialized in the final stages. This indicates that diffusion-based methods require significantly more time and sampling steps to converge, whereas flow matching efficiently establishes image structure much earlier.

\section{Limitations and Future Work}
PGFM is designed as a lightweight control layer on top of a frozen flow-matching generator, and its performance is therefore bounded by the quality and class-coverage of the underlying pretrained model and VAE latent space. While our early-stop and trust-region mechanisms improve stability, the method still involves a small set of hyperparameters (e.g., $\lambda$, $s_{\mathrm{end}}$, $\rho(t)$, and $\eta_{\mathrm{init}}$) whose optimal values can vary with dataset scale, resolution, and class count. Thus, developing more automatic, data-driven schedules remains an important direction. In addition, prototype discovery currently relies on simple $k$-means in latent space, which may be suboptimal for fine-grained classes or long-tailed settings; exploring stronger mode discovery (e.g., hierarchical clustering, density-based methods, or feature-space prototypes) is promising. Finally, our current control is prototype-centric and does not explicitly optimize downstream training objectives. So, our future work could integrate proxy training signals (e.g., gradient/influence estimates) into flow guidance, extend PGFM to larger backbones, and study theoretical guarantees on manifold preservation and mode coverage under controlled ODE dynamics.

\end{document}